%% file: main.tex
\documentclass[10pt,twocolumn,letterpaper]{article}

\usepackage[usenames,dvipsnames]{xcolor}
\usepackage{cvpr}
\usepackage{times}
\usepackage{epsfig}
\usepackage{subfig}
\usepackage{graphicx}
\usepackage{amsmath}
\usepackage{amssymb}
\usepackage{etoolbox}
\usepackage{refcount}
\usepackage{sidecap}
\usepackage{multirow}

\usepackage[breaklinks=true,bookmarks=false]{hyperref}

\cvprfinalcopy 

\DeclareMathOperator*{\argmax}{argmax}
\DeclareMathOperator*{\argmin}{argmin}

\def\cvprPaperID{352} 

\ifcvprfinal\fi
\begin{document}

\title{Inferring the Intentions of Actions in Images}
\title{Why are you doing that?}
\title{Inferring Human Motivation in Images}
\title{Inferring the Why in Images}
\title{Mining language to infer intention in images}
\title{Why did he do that?\\Inferring desires with common knowledge}
\title{Inferring Desires with Commonsense}
\title{Inferring people's motivation with commonsense mined from the web}
\title{Understanding motivations from images}
\title{Predicting the motivations behind actions}
\title{Predicting Motivations of Actions by Leveraging Text}

\author{
Carl Vondrick \; \; Deniz Oktay \; \; Hamed Pirsiavash$\dagger$ \; \; \; Antonio Torralba\\
Massachusetts Institute of Technology\; \;\hspace{1em} $\dagger$University of Maryland, Baltimore County \\
\texttt{\{vondrick,denizokt,torralba\}@mit.edu} \; \;  \texttt{hpirsiav@umbc.edu}
}


\newcommand{\fix}{\marginpar{FIX}}
\newcommand{\new}{\marginpar{NEW}}

\maketitle

\begin{abstract}
\input{abstract.tex}
\end{abstract}

\input{introduction.tex}

\input{related.tex}
\input{method.tex}
\input{evaluation.tex}
\input{conclusion.tex}

{\small
\textbf{Acknowledgements:} We thank Lavanya Sharan for important discussions,
Adria Recasens for help with the gaze dataset, and Kenneth Heafield and Christian
Buck for help with transferring the 6 TB language model across the Atlantic
ocean.
We gratefully acknowledge the support of NVIDIA Corporation with the donation of the Tesla K40 GPU used for this research.
This work was supported by NSF grant IIS-1524817, and by a Google
faculty research award to AT, and a Google PhD fellowship to CV.
}

{
\small
\bibliographystyle{ieee}
\bibliography{main}
}

\end{document}

%% file: abstract.tex
Understanding human actions is a key problem in computer vision. However,
recognizing actions is only the first step of understanding what a person is doing.
In this paper, we introduce the problem of predicting \emph{why} a person has
performed an action in images. This problem has many applications in human activity understanding,
such as anticipating or explaining an action. To study this problem,
we introduce a new dataset of people performing actions annotated with likely motivations.
However, the information in an image alone
may not be sufficient to automatically solve this task. Since humans can rely
on their lifetime of experiences to infer motivation, we propose to give computer
vision systems access to some of these experiences by using recently developed
natural language models to mine knowledge stored in massive amounts of text.
While we are still far away from fully understanding motivation, our
results suggest that transferring knowledge from language into vision can help
machines understand why people in images might be performing an action.

%% file: introduction.tex
\section{Introduction}




\begin{figure}[htbp]
\centering
\includegraphics[width=\linewidth]{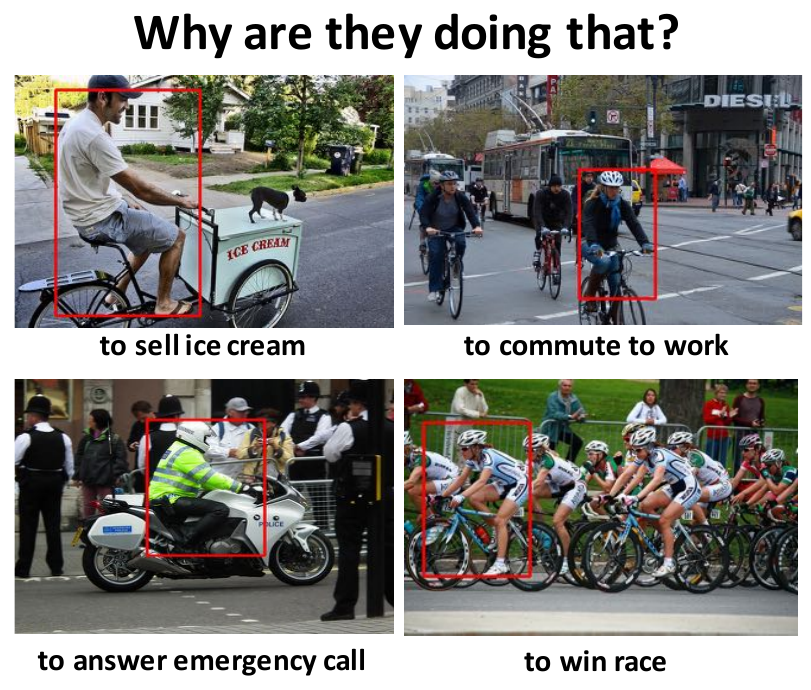}

\caption{\textbf{Understanding Motivations:} You can probably recognize that all of these people are riding bikes. Can you tell \emph{why} they are riding their bikes? In this paper, we learn to predict the motivations of people's actions by leveraging large amounts of text.}

\label{fig:teaser}
\end{figure}



Recognizing human actions is an important problem in computer vision.
However, recognizing actions is only the first step of understanding what a
person is doing. For example, you can probably tell that the people in Figure \ref{fig:teaser}
are riding bicycles. Can you determine \emph{why} they are riding bicycles? 
Unfortunately, while computer vision systems today can recognize
actions well, they do not yet understand the intentions and
motivations behind people's actions.

Humans can often infer why another person performs an action, in part
due to a cognitive skill known as the theory of mind \cite{wimmer1983beliefs}.
This capacity to infer another
person's intention may stem from the ability to impute our own beliefs onto others
\cite{baker2009action,saxe2003people}. For example, if we needed to commute to work, we might choose to ride our bicycle, similar to the top right in Figure \ref{fig:teaser}. Since we would be commuting to work in that situation, we might assume others in a similar situation would do the same. 

In this paper, we seek to predict the motivation behind people's
actions in images. To our knowledge,
inferring \emph{why} a person is performing an action from images has not
yet been extensively explored in computer vision. We believe that predicting motivations
can help understand human actions, such as anticipating or explaining an action.


To study this problem, we first assembled an image dataset of people (about $10,000$ people) and annotated
them with their actions, motivations, and scene.
We then combine these labels with state-of-the-art image
features \cite{zhou2014learning}  to train
classifiers that predict a person's motivation from images.  However,
visual features alone may not be sufficient to automatically solve
this task.  Humans can rely on a lifetime of experiences to predict motivations. 
How do we give computer vision systems access to similar experiences?

We propose to transfer knowledge from unlabeled text into visual classifiers in order to predict motivations.  Using large-scale language models \cite{heafield2011kenlm} estimated on billions
of web pages \cite{buck2014n}, we can acquire knowledge about
people's experiences, such as their interactions with objects, their
environments, and their motivations. We present an approach that integrates
these signals from text with computer vision to better infer motivations. While we are
still a long way from incorporating human experiences into a computer system, our
experiments suggest that we can predict
motivations with some success. By transferring knowledge acquired
from text into computer vision, our results suggest that we can predict why a
person is engaging in an action better than a simple vision only approach.

The primary contribution of this paper is introducing the problem of
predicting the motivations of actions to the computer vision community.
Since humans are able to reliably perform this task, we
believe that answering ``why'' for human actions is an interesting research problem to work on.
Moreover, predicting motivations has several applications in understanding and forecasting actions. 
Our second contribution is to use 
knowledge mined from text on the web to improve computer vision systems.  Our results
suggest that this knowledge transfer may be beneficial for predicting human motivation.  The
remainder of this paper describes this approach in detail. Section 2 first reviews related work.
Section 3 then introduces a new dataset for this task. Section 4 describes our model
that uses a factor graph composed of visual classifiers and pairwise potentials estimated from text. Section 5
presents experiments to analyze the approaches to predict motivation.

%% file: related.tex
\begin{figure*}
\includegraphics[width=\linewidth]{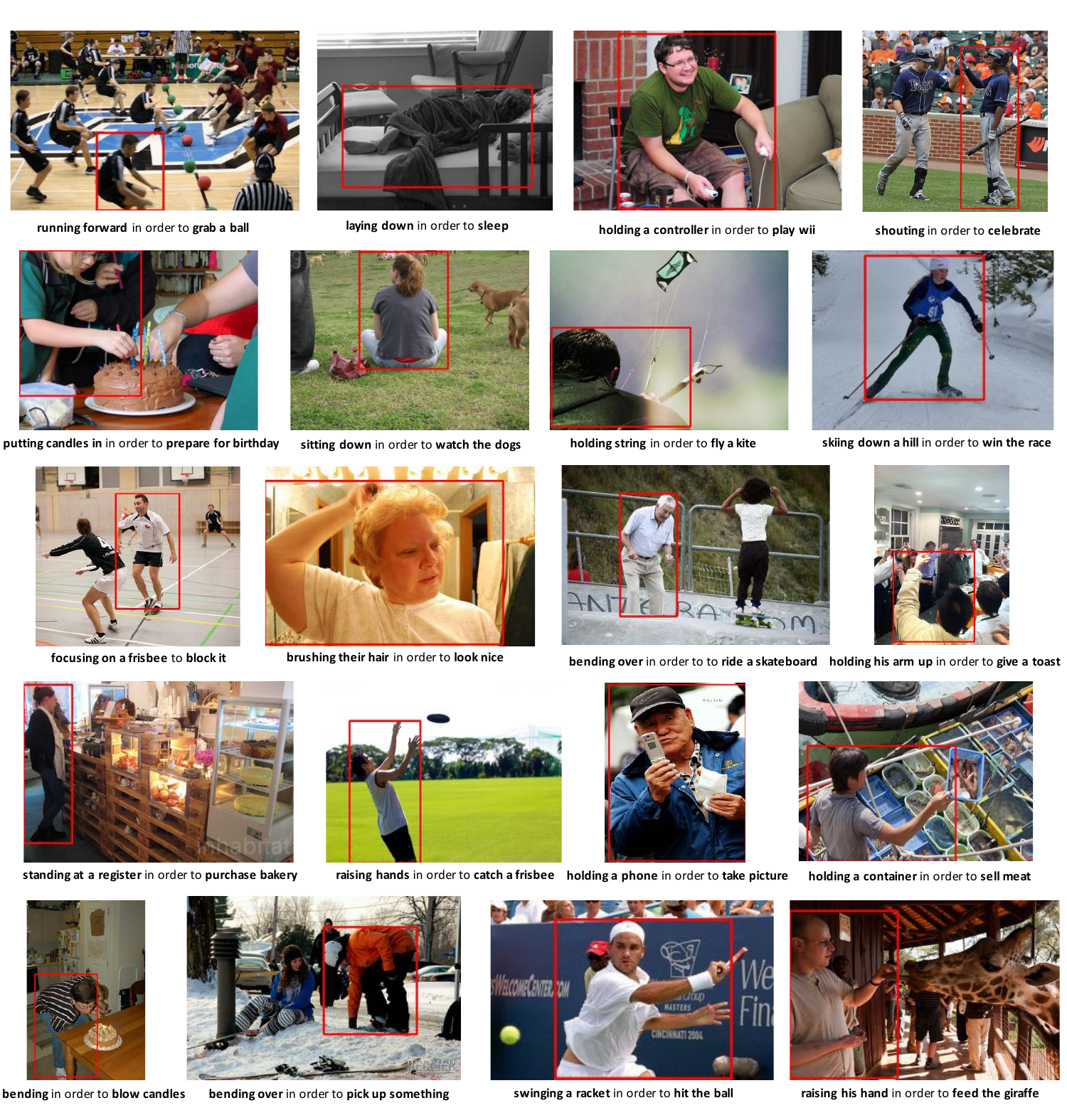}
\vspace{-2em}
\caption{\textbf{Motivations Dataset:} We show some example images, actions, and motivations from our dataset. Below each image we write a sentence in the form of ''\texttt{action} in order to \texttt{motivation}.'' We use this dataset to both train and evaluate models that predict people's motivations. The dataset consists of around $10,000$ people. Notice how the motivations are often outside the image, either in space or time.}
\label{fig:dataset}
\end{figure*}
\section{Related Work}

\textbf{Motivation in Vision:} Perhaps the most related to our paper is work
that predicts the persuasive motivation of the photographer who captured an
image \cite{persuasion}. However, our paper is different because we seek to
infer the motivation of the person \emph{inside} the image, and not the
motivation of the photographer. 

\textbf{Action Prediction:} There have been several works in robotics that
predicts a person's imminent next action from a sequence of images
\cite{song2013predicting,mcghan2012human,kelley2012deep,elfring2014learning,koppula2013anticipating}.
In contrast, we wish to deduce the motivation of actions in a single image, which may be related to what will happen next.  There also has been work in forecasting activities
\cite{kitani2012activity,patchtofuture}, inferring goals \cite{xie2013inferring}, and detecting early events
\cite{hoai2012max}, but they are interested in predicting the future in videos
while we wish to explain the motivations of actions of people in images. We believe insights into motivation can help further progress in action prediction.

\textbf{Action Recognition:} There is a large body of work studying how to recognize actions in images \cite{poppe2010survey,DBLP:journals/csur/AggarwalR11}. 
Our problem is related since in some cases the motivation can be seen as a high-level action. However, we are interested in understanding the motivation of the person engaging in an action rather than the recognizing the action itself. Our work complements action recognition because we seek to infer \emph{why} a person is performing an action.


\textbf{Commonsense Knowledge:} There are promising efforts in progress to acquire
commonsense sense for use in computer vision tasks
\cite{zitnick2013bringing,chen2013neil,divvalalearning,fouhey2014predicting,zhu2014reasoning}. In this paper, we also
seek to put commonsense knowledge into computer vision, but we instead attempt to
extract it from written language.

\begin{figure*}
\includegraphics[width=0.33\linewidth]{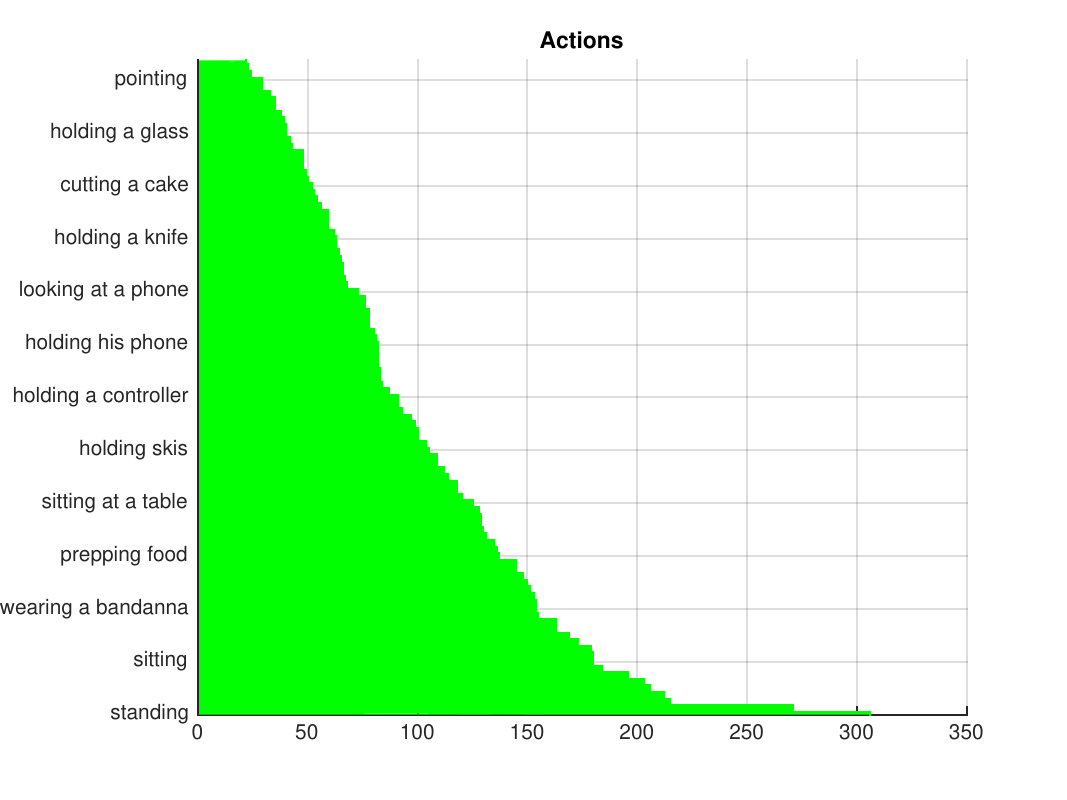}
\includegraphics[width=0.33\linewidth]{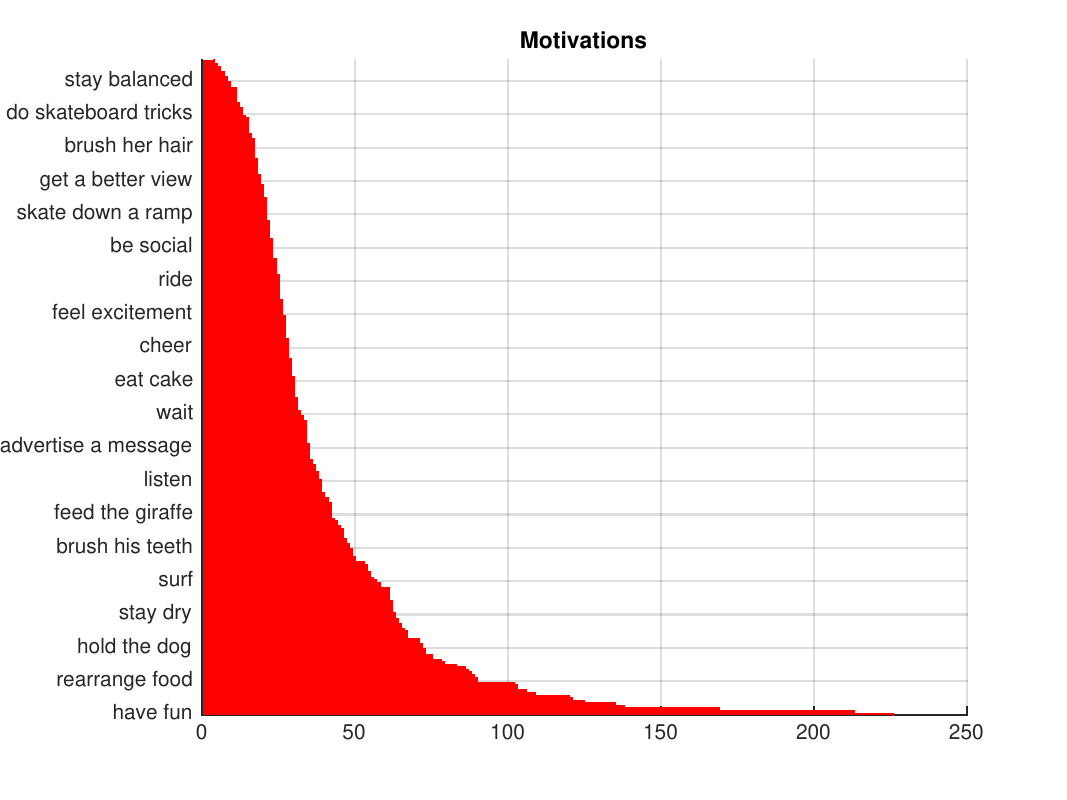}
\includegraphics[width=0.33\linewidth]{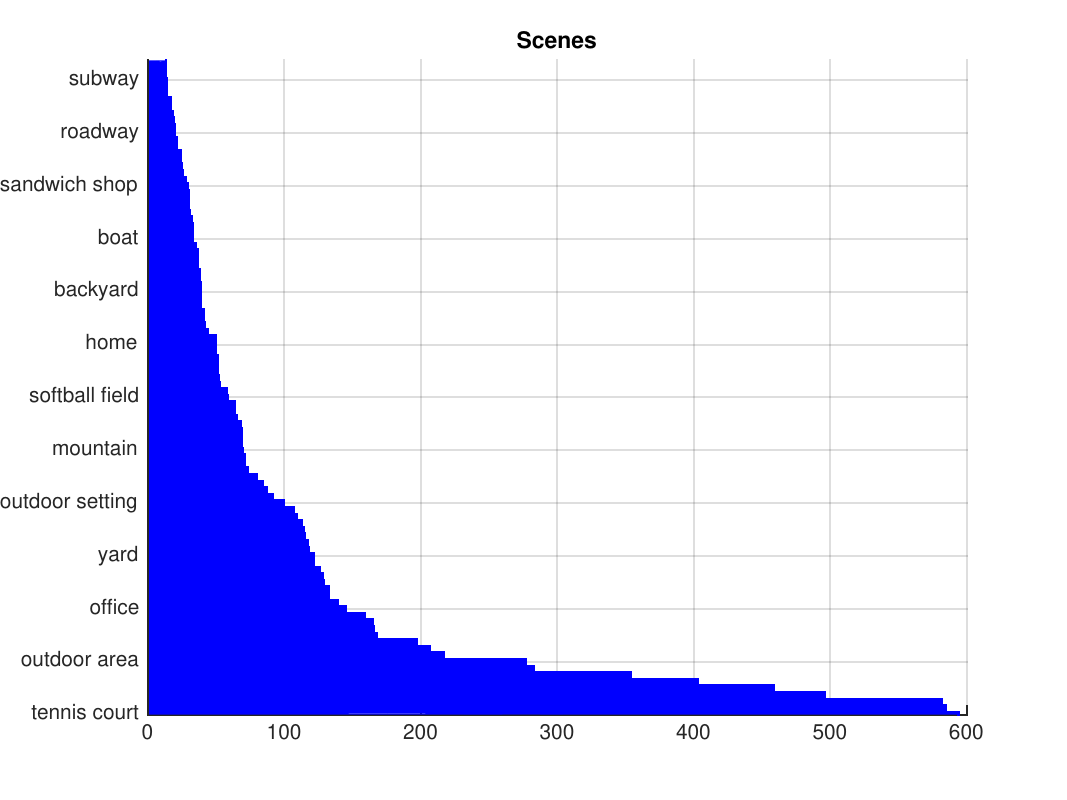}
\caption{\textbf{Statistics of Dataset:} We show a histogram of frequencies of the actions, motivations, and scenes in our dataset. There are $100$ actions, $256$ motivations, and $100$ scenes. Notice the class imbalance. On the vertical axis, not all categories are shown due to space restrictions.}
\label{fig:dataset-stats}
\end{figure*}

\textbf{Language in Vision:} The community has recently been incorporating natural
language into computer vision, such as
generating sentences from images \cite{kulkarni2011baby,karpathy2014deep,vinyals2014show}, producing visual
models from sentences \cite{zitnicklearning,wang2009learning}, and aiding in
contextual models \cite{patel2013language,Thu13}. In our work,
we seek to mine language models trained on a massive text corpus to extract 
some knowledge that can assist computer vision systems. 

\textbf{Visual Question Answering:} There have been several efforts to develop
visual question and answering systems in both images \cite{antol2015vqa,tapaswi2015movieqa} and  videos. One could
view answering why a person performs an action as a subset of the more general
visual QA problem. However, we believe understanding motivations is an
important subset to study specifically since there are many applications, such
as action forecasting. Moreover, our approach is different from most visual question answering
systems, as it jointly infers the actions with the motivations, and also provides a structured output
that more suitable for machine consumption.

%% file: method.tex
\section{Dataset}

On the surface, it may seem difficult to collect data for this task
because people's motivations are private and not directly observable.  However,
humans do have the
ability to think about other people's thinking
\cite{baker2009action,saxe2003people}. Consequently, we instruct 
crowdsourced workers to examine images of people and predict their
motivations, which we can use as both training and testing data.
%

\begin{figure}
\includegraphics[width=\linewidth]{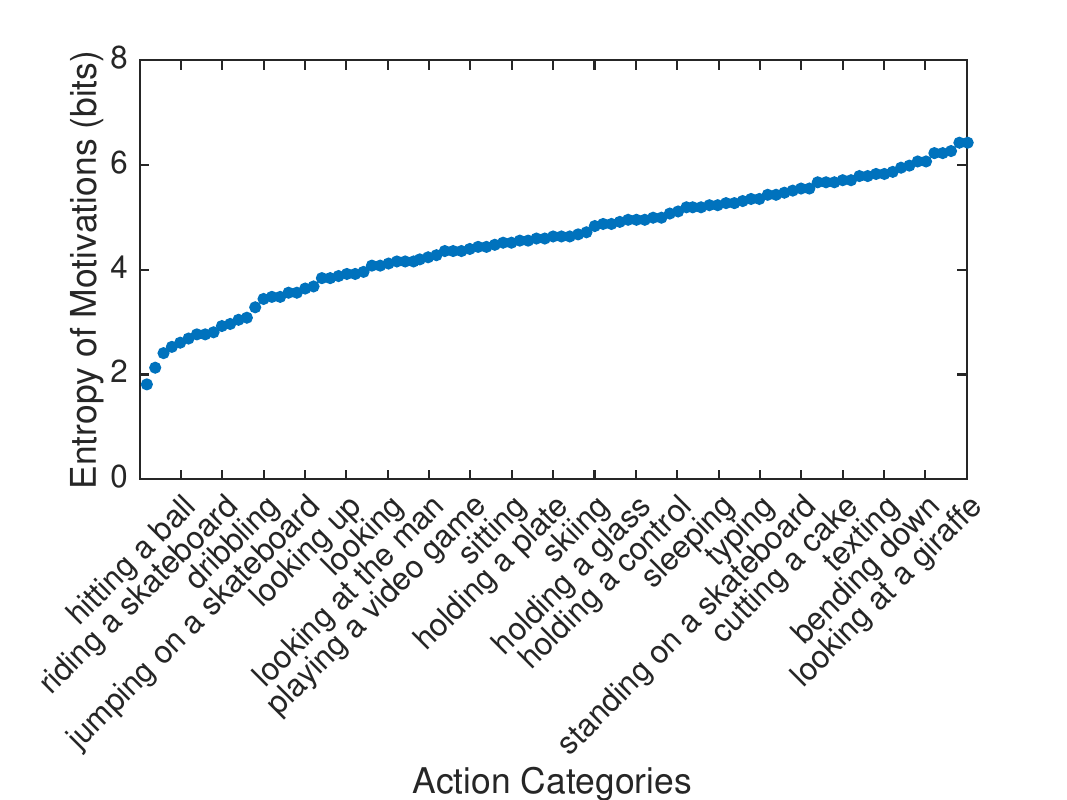}
\caption{\textbf{Are motivations predictable from just actions?}  We calculate the probability of a motivation conditioned on the action, and plot the entropy for each action. If motivations could be perfectly predicted from actions, the curve would be a straight line at the bottom of the graph (entropy would be $0$). If motivations were unpredictable from actions, the curve would be at the top (maximum entropy of $8$). This plot suggests that actions are correlated to the motivations, but it is not possible to predict the motivations only given the action. To predict motivations, we likely
need to reason about the full scene.}
\label{fig:predictable}
\end{figure}

\begin{figure}
\includegraphics[width=\linewidth]{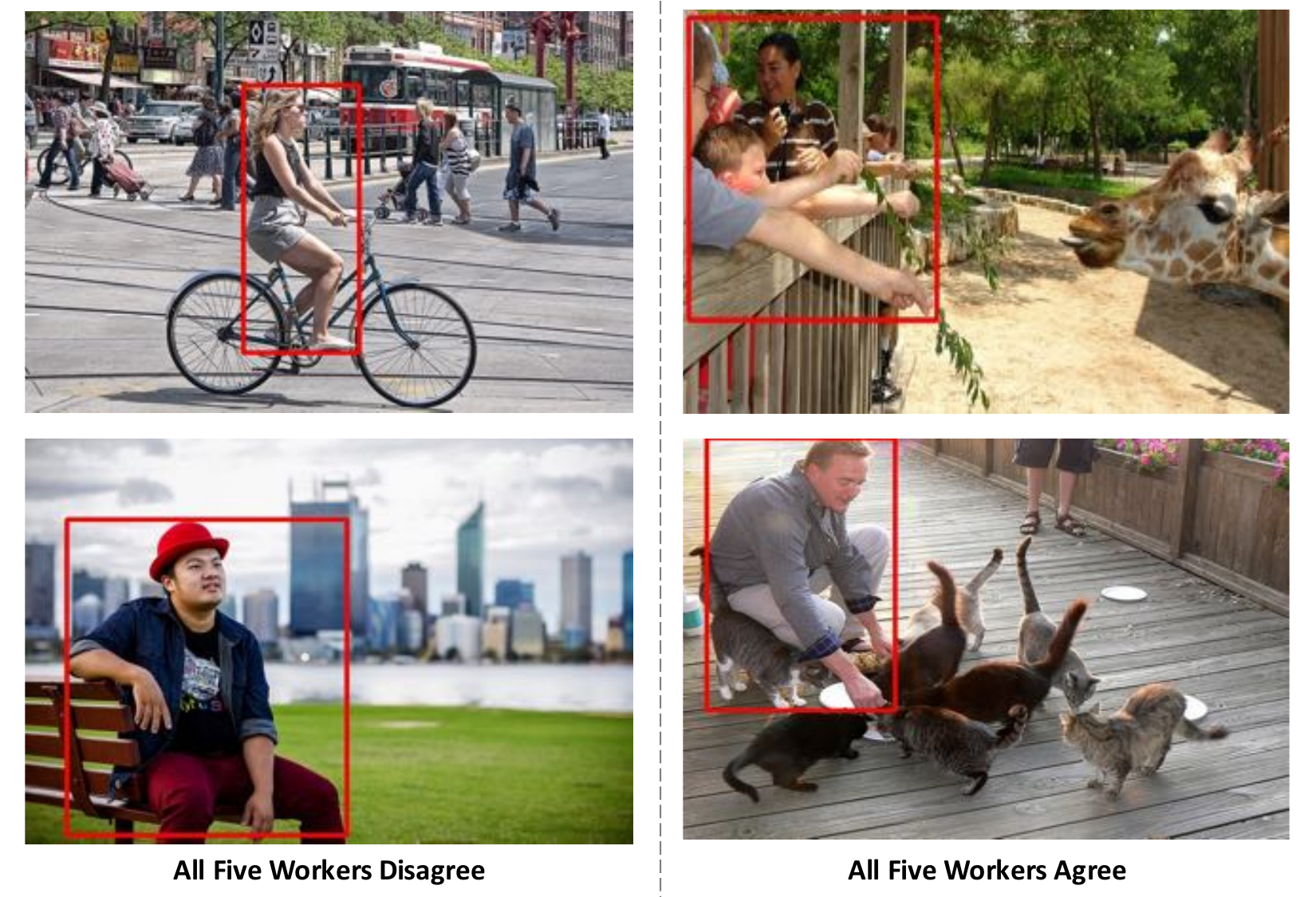}
\caption{\textbf{Which images have consistent motivations?} On the left, we show some images from our test set where all workers disagreed on the motivation. On the right, we show images where all workers agreed.}
\label{fig:agreement}
\end{figure}

We compiled a dataset of images of people by selecting $10,191$ people from
Microsoft COCO \cite{titlein2014microsoft}, and annotating motivations with
Mechanical Turk. 
In building this dataset, we found there were important choices for collecting
good annotations of motivations. We made sure that these images did not have any person
looking at the camera (using \cite{recasens2015they}), as otherwise the dominant motivation would be ``to take
photo.'' We wish to study natural motivations, and not ones where the person is
aware of the photographer.  We instructed workers on Amazon Mechanical Turk to
annotate each person with their current action, the scene, and their
motivation. We originally required workers to pick actions from a pre-defined
vocabulary, but we found this was too restrictive for workers. We had difficulty coming up
with a vocabulary of actions, possibly because the set of human actions may not
be well-defined. Consequently, we decided to allow workers to write short
phrases for each concept. Specifically, we had workers fill in the blanks for
two sentences: a) ``the person is \texttt{[type action]} in order to
\texttt{[type motivation]}'' and b) ``the person is in a \texttt{[type
scene]}.'' After data collection, we manually corrected the spelling.

We show examples from our dataset in Fig.\ref{fig:dataset}.  The images in the
dataset cover many different natural settings, such as indoor activities,
outdoor events, and sports. Since workers could type in any short phrase for
motivations, the motivations in our dataset vary. In general, the motivations
tend to be high-level activities that people do, such as ``celebrating'' or
``looking nice''.  Moreover, while the person's action is usually readily
visible, people's motivations are often outside of the image, either in space
or time. For example, many of the motivations have not happened yet, such as
raising one's hands because they want to catch a ball. 

\begin{table*}[tb]
\centering
\begin{tabular}{l | l}
Relationship & Query to Language Model \\
\hline
action & the person is \texttt{action} \\
motivation & the person wants to \texttt{motivation} \\
scene & the person is in a \texttt{scene} \\
action + motivation & the person is \texttt{action} in order to \texttt{motivation} \\
action + scene & the person is \texttt{action} in a \texttt{scene} \\
motivation + scene & the person wants to \texttt{motivation} in a \texttt{scene} \\
action + motivation + scene & the person is \texttt{action} in order to \texttt{motivation} in a \texttt{scene} \\
\end{tabular}

\caption{\textbf{Templates for Language Model:} We show examples of the queries we make to the language model. We combinatorially replaced \texttt{tokens} with words
from our vocabulary to score the relationships between concepts.}
\label{tab:queries}
\end{table*}

Since we instructed workers to type in simple phrases, workers frequently wrote
similar sentences. To merge these, we cluster each concept. We first embed each
concept into a feature space with skip-thoughts \cite{kiros2015skip}, and
cluster with $k$means. For actions and scenes, we found $k = 100$ to be
reasonable.  For motivations, we found $k = 256$ to be reasonable.
After clustering, we use the member in each cluster that is closest to the center
as the representative label for a cluster.
Fig.\ref{fig:dataset-stats} shows the distribution of motivations in our
dataset.  This class imbalance shows one challenge of predicting motivations
because we need to acquire knowledge for many categories. Since collecting such
knowledge manually with images (e.g.\ via annotation) would be expensive, we
believe language is a promising way to acquire some of this knowledge.

We are interested in analyzing the link between actions and motivations.
Can motivations be predicted from the action alone? To explore this, we 
calculate the distribution of motivations conditioned on an action, and plot the
entropy of these distributions in Fig.\ref{fig:predictable}. If motivations were predictable
given the action, then the entropy would be zero. On the other extreme, if motivations were
uncorrelated with actions, then the entropy would be maximum (i.e., $-\log_2\left(256\right) = 8$). Interestingly,
the motivations in our dataset lie between these two extremes, suggesting that motivations
are related to actions, but not the same.

Finally, we split the dataset into $75\%$ for training, and the rest for testing.
To check human consistency at this task, we annotated the test set $5$ times.
Two workers agreed on the motivation $65\%$ of the time, and three workers
agreed $20\%$ of the time. We compare this to the agreement if workers were to
annotate random motivations: two random labels agree $6\%$ of the time, and three random labels agree less than $1\%$ of the time. This suggests there is some structure in the data that
the learning algorithm can utilize. However, the problem may also emit
multi-modal solutions (people can have several motivations in an image).
We show example images where workers agree and disagree in Fig.\ref{fig:agreement}.

\section{Predicting Motivations} 

In this section, we present our approach to predict the motivations behind people's actions. 
We first describe a vision-only approach that estimates motivation from image features.
We then introduce our main approach that combines
knowledge from text with visual recognition to infer motivations.
\subsection{Vision Only Model}

Given an image $x$ and a person of interest $p$, a simple method can try to predict the motivation 
using only image features. Let $y \in \{1 \ldots M\}$ represent a possible motivation for
the person. We experimented with using a linear model to predict the most likely motivation: 
\begin{align}
\argmax_{y \in \{1, \ldots, M\}} \; w_{y}^T \phi(x, p)
\end{align}
where $w_y \in \mathbb{R}^D$ is a classifier that predicts the motivation $y$ from
image features $\phi(x) \in \mathbb{R}^D$. We can estimate $w_y$ by training an $M$-way linear classifier
on annotated motivations. We use one versus rest for multi-class classification. In our experiments, we use this model as a baseline.

\subsection{Extracting Commonsense from Text}

We seek to transfer some knowledge from text into the visual classifier to help predict motivation. 
Our main idea is to create a factor graph over several concepts (actions, motivations, and scenes).
The unary potentials come from visual classifiers, and the potentials for the relationships between
concepts can be estimated from large amounts of text. 

\begin{figure}[t]
\centering
\includegraphics[width=0.6\linewidth]{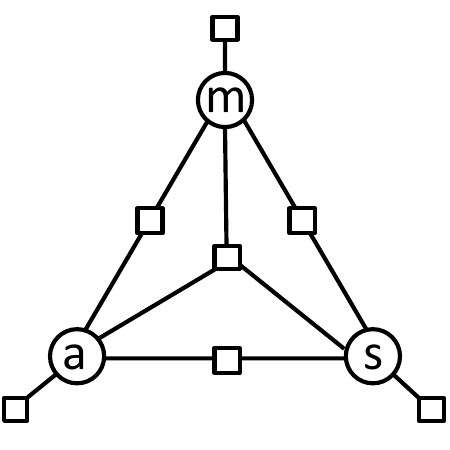}
\caption{\textbf{Factor Graph Relating Concepts:} We visualize the factor graph for our model. $a$ refers to action, $s$ for scene,
and $m$ for motivation. We use language to estimate potentials.}
\label{fig:factor}
\end{figure}

Let $x$ be an image, $p$ be a person in the image, and 
$y_i \in \{1 \ldots k_i\}$ be its corresponding labels for $1 \le i \le K$. In our case, $K=3$ because each image is annotated with a scene, action, and motivation.
We score a possible labeling configuration $y$ of concepts with the function:
\begin{equation}
\begin{aligned}
&\Omega(y | x, p ; w, u) = \sum_{i}^K w_{y_i}^T \phi_i(x,p)  + \sum_{i}^K u_i L_i(y_i) \\
&+ \sum_{i < j}^K u_{ij} L_{ij}(y_i, y_j) + \sum_{i < j < k}^K u_{ijk} L_{ijk}(y_i, y_j, y_k)
\end{aligned}
\label{eqn:objective}
\end{equation}
where $w_{y_i} \in \mathbb{R}^{D_i}$ is the unary term for the concept $y_i$ under visual features $\phi_i(\cdot)$, and $L(y_i, y_j, y_k)$ are potentials
that scores the relationship between the visual concepts $y_i$, $y_j$, and $y_k$. The terms
$u_{ijk} \in \mathbb{R}$ calibrate these potentials with the visual classifiers.  We will learn both $w$ and $u$, while $L$ is estimated from text. 
Our model forms a third order factor graph, which we visualize in Fig.\ref{fig:factor}.

\begin{table}
\begin{tabular}{l | l l}
& Action & Motivation \\
\hline
\multirow{ 5}{*}{High Scoring} & watching & see \\
& reading & learn \\
& talking & listen \\
& talking & learn \\
& running & play \\
& $\cdots$ & $\cdots$ \\
\multirow{ 5}{*}{Low Scoring} & watching & type on laptop \\
& skiing & look at truck \\
& sleeping & see a giraffe  \\
& reading & cut wedding cake \\
& riding skateboard & get cake \\
& $\cdots$ & $\cdots$ \\
\end{tabular}
\caption{\textbf{Example Language Potentials:} By mining billions of web-pages, we can extract
some knowledge about the world. This table shows some pairs of concepts, sorted by the score
from the language model.}
\label{fig:example-lang}
\end{table}

In order to learn about the relationships between concepts, we mine large
amounts of text.  Recent progress in natural language processing has created
large-scale language models that are trained on billions of web-pages
\cite{buck2014n,heafield2011kenlm}. These models work by ultimately calculating
a probability that a sentence or phrase would exist in the training corpus.
Since people usually do not write about scenarios that are rare
or impossible, we can query these language models to score the relationship
between concepts. Fig.\ref{fig:example-lang} shows some pairs of actions and motivations, sorted by the score from the language model. For example, the language model that we use predicts that
``reading in order to learn'' is more likely than ``reading in order to cut wedding cake,'' likely because stories about people reading to cut wedding cake is uncommon. 

Specifically, to estimate $L(\cdot)$ we ``fill in the blanks'' for sentence templates.
Tab.\ref{tab:queries} shows some of the templates we use. For example, to score
the relationships between different motivations and actions, we query the language model
for ``the person is \texttt{action} in order to \texttt{motivation}'' where \texttt{action} and \texttt{motivation}
are replaced with different actions and motivations from our dataset. Since querying is automatic, we can efficiently
do this for all combinatoric pairs. In the most extreme case, we query for tertiary terms for all
possible combinations of motivations, actions, and scenes. In our experiments, we use a $5$-gram language model 
that outputs the log-probabilities of each sentence.


Note that, although ideally the unary and binary potentials would be redundant
with the ternary language potentials, we found including the binary potentials
and learning a weight $u$ for each improved results.  We believe this is the
case because the binary language model potentials are not true marginals of the
ternary potentials as they are built by a limited number of queries. Moreover,
by learning extra weights, we increase the flexibility of our model, so we can
weakly adapt the language model to our training data.

\subsection{Inference}

Joint prediction of all concepts including motivation corresponds to calculating the most likely
configuration $y$ given an image $x$ and learned parameters $w$ and $u$ over the factor graph:
\begin{align}
y^* = \argmax_y \; \Omega(y | x, p; w, u)
\end{align}
We often require the $n$-best solutions, which can
be done efficiently with approximate approaches such as $n$-best MAP estimation
\cite{batra2012diverse,lawler1972procedure} or sampling techniques \cite{porway2011c,grante}.
However, we found that, in our experiments, it was tractable to evaluate all configurations with a simple matrix multiplication,
which gave us the exact $n$-best solutions in a few seconds on a high-memory server.

\subsection{Learning}

\begin{table*}[tb]
\centering
\begin{tabular}{l l | c c c}
  & & \multicolumn{3}{|c}{Median Rank} \\
  & Method & Motivation & Action & Scene \\
  \hline
  \multirow{ 6}{*}{Automatic} & Random & 62  & 29  & 15 \\
  & Vision & 39 & 17 & 4 \\
  & Vision + Person & 42 & 18  & 4 \\
  & Vision + Text & 28  & 15 & 3  \\
  & Vision + Person + Text  & \textbf{27} & 14 & 3 \\
  & Vision + Person + Text (Wikipedia) & 28 & 14 & 3 \\
  \hline
  \multirow{ 3}{*}{Diagnostics} 
  & Vision + Person + Text + Action & 26 & - & 3 \\
  & Vision + Person + Text + Scene & 27 & 14 & -  \\
  & Vision + Person + Text + Action + Scene & 25  & - & - \\
\end{tabular}
\caption{\textbf{Evaluation of Median Rank:} We show the median rank of the ground truth motivations
in the predicted motivations, comparing several methods. Lower is better with
$1$ being perfect. There are $256$ motivations, $100$ actions, and $100$
scenes. In the bottom of the table, we show diagnostic experiments, where we give the
classifier access to the ground truth of other concepts, and evaluate how well it can infer the remaining concepts.}
\label{tab:results}
\end{table*}

We wish to learn the parameters $w$ for the visual features and $u$ for the language potentials
using training data of images and their corresponding labels, $\{x^n, y^n\}$. 
Since our scoring function in Eqn.\ref{eqn:objective} is linear on the model parameters $\theta = [w; u]$, we can write the scoring function in the linear form $\Omega(y | x, p, w, u) = \theta^T \psi(y, x, p)$. We want to learn $\theta$ such that the labels matching the ground truth score higher than incorrect labels. We adopt a max-margin structured prediction objective:
\begin{equation}
\begin{aligned}
&\argmin_{\theta, \xi^n \ge 0} \; \frac{1}{2} ||\theta||^2 + C \sum_{n} \xi^n \quad \textrm{s.t.} \\ &\theta^T \psi(y^n, x^n, p^n) - \theta^T \psi(h, x^n, p^n) \ge 1 - \xi^n \; \forall_n, \forall_{h \ne y^n}
\end{aligned}
\label{eqn:svm}
\end{equation}
The linear constraints state that the score for the correct label $y^n$ should be larger than that of any other hypothesized label $h^n$ by at least $1$. We use a standard 0-1 loss function that incurs a penalty if any of the concepts do not match the ground truth. This optimization is equivalent to a structured SVM and can be solved by efficient off-the-shelf solvers \cite{joachims2009cutting,ramanan2013dual}. Note that one could use hard-negative mining to solve this optimization problem for learning. However, in our experiments, it did not improve results, possibly because we train the model on a high-memory machine.

\subsection{Indicating Person of Interest}

Finally, to predict somebody's motivation, we must specify the person of
interest. To do this, we crop the person, and extract visual features for the
close-up crop. We then concatenate these features with the full image features.
We use this concatenation to form $\phi(x,p)$.

%% file: evaluation.tex
\section{Experiments}

\begin{figure*}[tb]
\centering
\includegraphics[width=\linewidth]{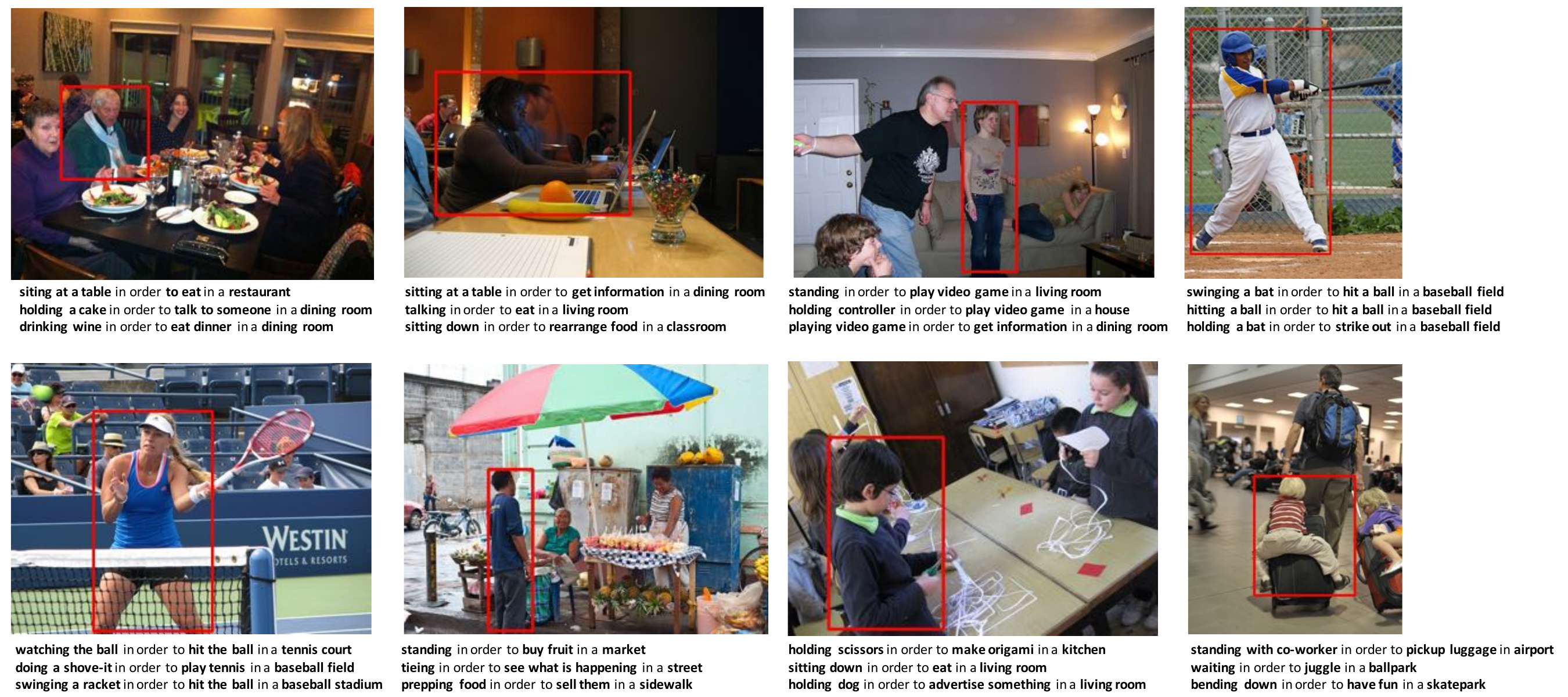}

\caption{\textbf{Example Results:} We show the top three predictions for some people in our evaluation set. The sentences are shown only to visualize results: we do not generate free-form captions. Even when the model is wrong, it usually predicts concepts that have plausible relations to other, which is possible because of the language model.}

\label{fig:qual-us}
\end{figure*}

In this section, we evaluate the performance at inferring motivations in images
from our models.
We first describe our evaluation setup, then we present our results.

\subsection{Experimental Setup}

We designed our experiments to evaluate how well we can predict the motivation
of people from our dataset. We assumed the person-of-interest is specified
since the focus of this work is not person detection. We computed features from
the penultimate layer ($\texttt{fc7}$) in the VGG-16 convolutional neural
network trained on Places \cite{zhou2014learning,simonyan2014very} due to their
strong performance on other visual recognition problems
\cite{razavian2014cnn,donahue2013decaf}.  We concatenate the features for both
the full image and a close-up crop of the person of interest (giving a $8,192$
dimensional feature vector). We experimented with a few different $C$ values,
and report the best performing ($0.001$). To compute the chance performance, we evaluate median
rank against a list sorted by the frequency of motivations.


%
%
%

\subsection{Evaluation}

We evaluate our approach on an image by finding the median rank of a ground truth
motivation in the max-marginals on all states for the motivation concept. This
is equivalent to the rank of ground truth motivation in the list of motivation
categories, sorted by their best score among all possible configurations. We 
use median rank because motivation prediction may have multi-modal solutions, and
we seek an evaluation metric that allows for multiple predictions. We
report the median rank of our full approach and the baseline in
Tab.\ref{tab:results}.

Our results suggest that incorporating knowledge from text can improve
performance at predicting motivations, compared to a vision-only approach.
Our interpretation is that text acts as a regularizer that helps the model perform
better on novel images during inference time. The baseline appears prone to over-fitting,
since the baseline performs slightly worse when we specify the person.
Moreover, our approach is significantly better than chance, suggesting that the
learning algorithm is able to use the structure in the data to infer
motivations. As one might expect, performance for our model improves when we specify the
person of interest in the image.

\subsection{Diagnostics}

For diagnostic purposes, the bottom of Tab.\ref{tab:results} shows the
performance of our approach if we had ideal recognition
systems for each visual concept. To give ideal detectors to our system, we can
constrain the unary potentials for the corresponding concept to the ground
truth.  Our
results suggest that if we had perfect vision systems for actions and
scenes, then our model would only slightly improve, and it would still not
solve the problem. This suggests that motivations are not the same as scenes or
actions. Rather, in order to improve performance further, we hypothesize
integrating additional visual cues such as human gaze \cite{recasens2015they}
and clothing will yield better results, motivating work in high-level visual
recognition.

To evaluate the impact of the language model, we also experimented with using
an alternative language model trained on text from Wikipedia instead of the
web. Here, the language model is estimated with much less data.
Interestingly, using the Wikipedia language model performs slightly worse at predicting motivations (by one point),
suggesting that a) leveraging more unlabeled text may help, and b) better language models can help computer vision tasks. We
believe that advances in natural language processing can help
computer vision systems recognize concepts, especially higher-level concepts
such as motivations.



We show a few examples of successes and failures for our approach in
Fig.\ref{fig:qual-us}. We hypothesize that our model often produces more
sensible failures because it leverages some of the knowledge available in
unlabeled text. 


%

%% file: conclusion.tex
\section{Conclusion}

While computers can recognize the actions of people in images with good performance, predicting the motivations
behind actions is a relatively less explored problem in computer vision. This problem is
important both for developing a full understanding of actions, as well as in applications, such
as anticipating future actions. 
We have released a new dataset to study this problem, and we have investigated how to 
transfer some knowledge from text to help predict motivations.

Our experiments indicate that there is still significant room for improvement. We suspect that advances in high-level
visual recognition can help this task. However, our results suggest that visual
information alone may not be sufficient for this challenging task.  We
hypothesize that incorporating common knowledge from other sources can help,
and our results imply that written language is one valuable source.
Our framework only transfers some of the commonsense knowledge into
vision models, and we hypothesize that continued work in commonsense reasoning for vision will help machines infer motivations.  We believe
that progress in natural language processing is one way to advance such high-level reasoning
in computer vision.